\documentclass[letterpaper]{article} % DO NOT CHANGE THIS
\usepackage{aaai25}  % DO NOT CHANGE THIS
\usepackage{times}  % DO NOT CHANGE THIS
\usepackage{helvet}  % DO NOT CHANGE THIS
\usepackage{courier}  % DO NOT CHANGE THIS
\usepackage[hyphens]{url}  % DO NOT CHANGE THIS
\usepackage{graphicx} % DO NOT CHANGE THIS
\usepackage{natbib}  % DO NOT CHANGE THIS AND DO NOT ADD ANY OPTIONS TO IT
\usepackage{caption} % DO NOT CHANGE THIS AND DO NOT ADD ANY OPTIONS TO IT
\usepackage{algorithm}
\usepackage{algorithmic}

\usepackage{newfloat}
\usepackage{listings}
\DeclareCaptionStyle{ruled}{labelfont=normalfont,labelsep=colon,strut=off} % DO NOT CHANGE THIS
\floatstyle{ruled}
\newfloat{listing}{tb}{lst}{}
\floatname{listing}{Listing}
\title{The Phantom of the Elytra - Phylogenetic Trait Extraction from Images of Rove Beetles Using Deep Learning - Is the Mask Enough?}
\author {
    Roberta Hunt\textsuperscript{\rm 1},
    Kim Steenstrup Pedersen\textsuperscript{\rm 1,\rm 2}
}
\affiliations {
    \textsuperscript{\rm 1}Department of Computer Science, University of Copenhagen\\
    \textsuperscript{\rm 2}Natural History Museum of Denmark\\
    r.hunt@di.ku.dk, kimstp@di.ku.dk
}

\usepackage{xcolor}

\begin{document}

\maketitle

\begin{abstract}
Phylogenetic analysis traditionally relies on labor-intensive manual extraction of morphological traits, limiting its scalability for large datasets. Recent advances in deep learning offer the potential to automate this process, but the effectiveness of different morphological representations for phylogenetic trait extraction remains poorly understood. In this study, we compare the performance of deep learning models using three distinct morphological representations — full segmentations, binary masks, and Fourier descriptors of beetle outlines. We test this on the Rove-Tree-11 dataset, a curated collection of images from 215 rove beetle species. Our results demonstrate that the mask-based model outperformed the others, achieving a normalized Align Score of $0.33\pm0.02$ on the test set, compared to $0.45\pm0.01$ for the Fourier-based model and $0.39\pm0.07$ for the segmentation-based model. The performance of the mask-based model likely reflects its ability to capture shape features while taking advantage of the depth and capacity of the ResNet50 architecture. These results also indicate that dorsal textural features, at least in this group of beetles, may be of lowered phylogenetic relevance, though further investigation is necessary to confirm this. In contrast, the Fourier-based model suffered from reduced capacity and occasional inaccuracies in outline approximations, particularly in fine structures like legs. These findings highlight the importance of selecting appropriate morphological representations for automated phylogenetic studies and the need for further research into explainability in automatic morphological trait extraction.
\end{abstract}

% Uncomment the following to link to your code, datasets, an extended version or similar.
%
%\begin{links}
%     \link{Datasets}{https://erda.ku.dk/archives/118d9022feb67f8eb7f7bc8bce71187f/published-archive.html}
%\end{links}

\section{Introduction}
% Why is phylogenetics important, why are we interested in removing texture,

Phylogenetic analysis is a cornerstone of evolutionary biology, providing insights into the relationships between species and their evolutionary histories. Traditionally, the reconstruction of phylogenies relies on genetic data, morphological traits, or a combination of both \cite{huelsenbeck1996combining}. However, extracting morphological traits for phylogenetic studies is often labor-intensive, requiring expert knowledge and manual annotation. Recent advances in deep learning offer a promising avenue to automate the extraction of morphological features from images, potentially revolutionizing the integration of morphology into phylogenetics \cite{hofmann2024inferring}.

Textural bias in deep learning models is a known issue \cite{hermann2020origins}, and as yet research has not explored how this presents itself in deep learning derived morphological traits.

With their diverse morphologies and extensive species richness, rove beetles provide a challenging but rewarding test case for automated phylogenetic trait extraction. However, a critical question remains: which representation of a beetle’s morphology is most effective for deep learning models to capture phylogenetically informative traits?

This study focuses on comparing the efficacy of three distinct approaches for deep learning derived morphological trait extraction: 1) masks of the beetle’s body, which provide a binary segmentation of the beetle’s silhouette; 2) Descriptors of Fourier epicycles derived from the beetle's outline, which capture its shape in a mathematically compact form \cite{lestrel1997fourier}; and 3) the full segmentation of the beetle body from a dorsal view including color and external structure. Each approach offers unique strengths in representing morphology: masks and fourier descriptors emphasize overall shape in two distinct ways and the segmentations provide textural, external structural and color information.

By evaluating these representations in terms of their ability to predict phylogenetic traits using deep metric learning, this study seeks to investigate how useful shape compared to textural information is to phylogenetic trait extraction.

\begin{figure*}[t!]
\centering
\includegraphics[width=.3\textwidth]{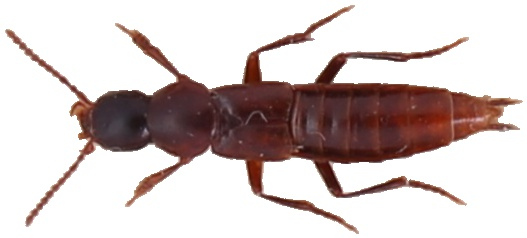}\hfill
\includegraphics[width=.3\textwidth]{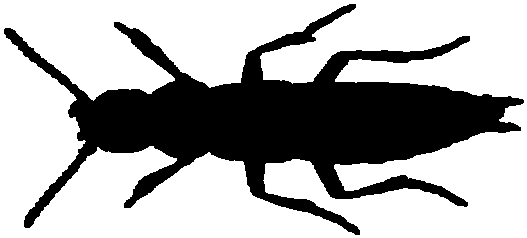}\hfill
\includegraphics[width=.3\textwidth]{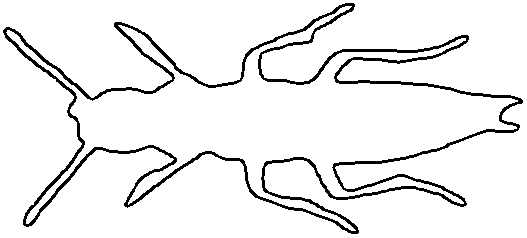}

\caption{Example input images. (left) original segmentation from dataset. (middle) binary mask extracted from segmentation. White is used as the background here to keep padding equivalent during data augmentations. (right) Outline produced using 200 Fourier descriptors.}
\label{fig:examples}
\end{figure*}

\section{Background}
% Morphological trait extraction, deep learning applied to ecology, 

% need to mention the newer models - ARtTree phyloGAN graphGAN, argue that this intermediate step lets us mimic biologists and we believe might be more interpretable in the future. adaptation of those methods to images is outside the scope of this paper, but would be very interesting to see. 
The extraction of phylogenetic traits from morphological data has been a central challenge in evolutionary biology. Prior to genetic analysis, it was the cornerstone of initial investigations of phylogenetic relationships. Since genetic data is becoming more accessible, there is an argument to be made for removing the focus on morphological traits entirely. However, the use of morphological traits remains common \cite{lee2015morphological}, partly because genetic data cannot always be extracted from older specimens, such as fossils or pinned insect collections, which form a significant foundation for our current understanding of the tree of life.

Traditionally, morphological analyses rely on manual measurements and qualitative descriptions of traits, requiring significant expertise and time. While these methods have provided valuable insights into evolutionary relationships, they are often subjective and difficult to scale for large datasets, such as those involving highly diverse taxa like rove beetles. With the rise of image-based analysis and computational methods, automated approaches to morphological trait extraction using AI are increasingly being explored to overcome these limitations \cite{hoyal2019deep, hunt2022rove, tsutsumi2023deep, hofmann2024inferring}.

Deep learning has shown remarkable success in automating feature extraction and classification tasks across various domains, including image recognition, medical imaging, and biodiversity monitoring. In evolutionary biology, deep learning models have recently been applied to tasks such as species identification and behavioural analysis. We refer to \cite{borowiec2022deep} for an overview of the current state of deep learning applied to ecological problems. 

Deep learning has also been applied to phylogenetics, with the emphasis on genetics. Recent methods for deep learning based inference have shown success on molecular data. Of particular note we mention phyloGFN \cite{zhou2023phylogfn} and ARTree \cite{xie2024artree}. These methods directly infer the phylogenetic trees from the molecular data without first generating trait matrices and have been shown to be competitive with Bayesian and parsimony methods. While we are excited by these methods, and recognize that they could be applied to morphological images, we see value from an explainability and adoptability side in investigating trait extraction as an intermediary step. We refer to \cite{mo2024applications} for a more detailed survey of deep learning phylogenetic methods applied to molecular data.

In the context of morphology, several approaches have been proposed to represent and analyze shapes. Among these, binary masks have been employed to capture the overall shape, providing a straightforward representation that highlights total morphological differences. Fourier descriptors have long been used in morphometrics to quantify shape variation \cite{lestrel1997fourier}, particularly in structures with smooth outlines, such as forensic anthropology \cite{caple2017elliptical}. By transforming shapes into a series of mathematical descriptors, Fourier descriptors reduces high-dimensional shape data into compact representations while retaining key geometric features. Previous studies have demonstrated the utility of each of these approaches \cite{shi2024few}. However, as far as we know this is the first attempt at comparing the performance of these representations for phylogenetic trait extraction using deep learning. 

This work contributes to the growing interdisciplinary field of deep learning and evolutionary biology by comparing morphological representations for phylogenetic analysis. It also highlights the potential of automated approaches to accelerate and improve the integration of morphology into large-scale phylogenetic studies.

\section{Dataset}
The Rove-Tree-11 dataset \cite{hunt2022rove} is a curated collection of over 13,000 segmented dorsal images of rove beetles (family Staphylinidae) and an associated 11-depth phylogeny specifically designed for the study of phylogenetic trait extraction using deep learning. This dataset includes images of 215 species of rove beetles from the collections at the Natural History Museum of Denmark and spans three sub-families: Staphylininae, Paederinae and Xantholinae, each respectively included as the train, validation and test set. Using separate sub-families for the train, validation, and test sets, rather than a typical stratified dataset as used in classification, allows us to evaluate the model's ability to generalize to completely unseen data. For more details about the dataset distribution and comparison with other datasets, we refer to the original paper \cite{hunt2022rove}.

\section{Methodology}
To evaluate the performance of different morphological representations for extracting phylogenetic traits, we include three distinct representations of rove beetle morphology: binary masks, Fourier descriptors, and the color pixel segmentations. Below, we detail our the data preprocessing, model architectures, and experimental procedures. Examples of each can be found in figure \ref{fig:examples}.

\subsection{Data Preprocessing}
The Rove-Tree-11 dataset includes segmented images and a phylogeny. Binary masks and Fourier descriptors of the outlines were obtained as follows:

\subsubsection{Binary Masks}
Since the dataset includes fully segmented images on a white background, binary masks can easily be obtained. To do this we blurred the images with a 3x3 kernel and thresholded them at 250. The blurring and thresholding help smooth the images to remove artifacts from jpeg compression.  

\subsubsection{Fourier Descriptors}
\begin{figure*}[t!]
\centering
\includegraphics[width=.22\textwidth]{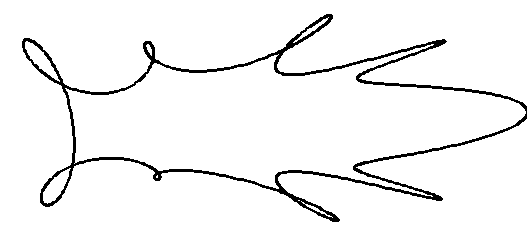}\hfill
\includegraphics[width=.22\textwidth]{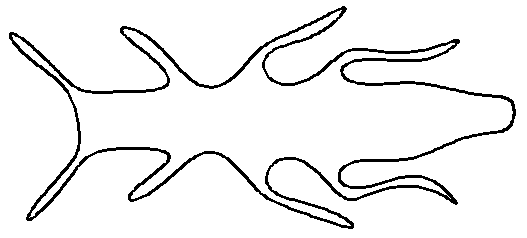}\hfill
\includegraphics[width=.22\textwidth]{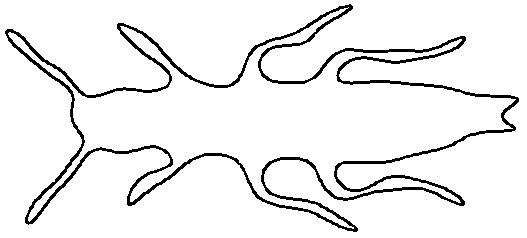}\hfill
\includegraphics[width=.22\textwidth]{images/fourier_200.png}

\caption{Example outlines produced using varying numbers of Fourier coefficients. From left to right: 20, 50, 100 and 200 Fourier coefficients used.}
\label{fig:fouriercoeffs}
\end{figure*}

The beetle’s outline was extracted from the binary mask and transformed into a set of Fourier descriptors using the Fourier Epicycles method \cite{lestrel1997fourier}. For each beetle we captured 200 Fourier coefficients with 200 associated angular velocities. Including the constant coefficient, this amounts to an input vector of length 402. Using 200 coefficients ensured that sufficient shape detail was preserved (see figure \ref{fig:fouriercoeffs}).

\subsection{Model Architectures}

\begin{table}[t]
\centering
%\resizebox{.95\columnwidth}{!}{
\begin{tabular}{r|r}
    {\bf Architecture} & {\bf No. Parameters}\\
    \hline
    ResNet50 & 23,770,304 \\
    DSeqEnc & 103,028 \\
\end{tabular}
\caption{Number of parameters of architectures used in this work}
\label{tab1:params}
\end{table}

For the pixel segmentations and the binary masks we used the architecture (ResNet50) and methodology in the Rove-Tree-11 paper with a few modifications \cite{hunt2022rove}. To make the training process more efficient we only trained for 50 epochs, and in order to use smaller GPUs we used gradient accumulation with a mini-batch size of 8 and 14 gradient accumulation steps to mimic the batch size of 112 from the Rove-Tree-11 paper. As opposed to the Rove-Tree-11 paper, only triplet loss was considered as it provided reasonable results and is one of the better known deep metric learning methods.

For the Fourier descriptors we used a simple encoder architecture (we call DSeqEnc) inspired by the DeepSequence model \cite{riesselman2017deep}. The original DeepSequence model is a simple autoencoder network shown to extract mutation and simple phylogenetic relationships from protein data. Since we have fewer input dimensions than the original paper, we decreased the size of the encoder layers, and since we are only interested in the latent representations we removed the decoder. Our architecture has 2 hidden layers of sizes 402x150 and 150x150, each followed by batch normalization and a ReLU function, and a final layer 150x128. Loss function and training regime are the same as the ResNet50 model. The latent size of 128 was chosen to match the Rove-Tree-11 paper and ResNet50 model embedding layer.

The number of parameters for each architecture can be found in table \ref{tab1:params}.

\subsection{Evaluation Metrics}
Following the evaluation approach in the Rove-Tree-11 dataset paper, phylogenies are estimated based on the extracted traits. The Align score \cite{nye2006novel}, now normalized (nAS) and normalized Robinson Foulds (nRF) scores are used to evaluate the estimated phylogenies against the ground truth phylogeny. An comparison of tree metrics is provided in \cite{kuhner2015practical} for reference.

We used the repository for the Rove-Tree-11 dataset paper as the code base, which is publicly available \cite{hunt2022rove}. Additional code for generating Fourier descriptors and the DSeqEnc Model is available from the authors upon request.

\section{Results}

\begin{table}[t]
\centering
%\resizebox{.95\columnwidth}{!}{
\begin{tabular}{l|l|l|l}
    {\bf Dataset} & {\bf Arch} & {\bf nAS $(\downarrow)$} & {\bf nRF $(\downarrow)$} \\ \hline
    Random & - & $0.66 \pm 0.04$ & $0.99 \pm 0.04$ \\
    Fourier & DSeqEnc & 0.45$\pm$0.01 & 0.93$\pm$0.00 \\
    Segmentations & ResNet50 & \underline{0.39$\pm$0.07}  &  \underline{0.89$\pm$0.05} \\
    Masks & ResNet50 & \textbf{0.33$\pm$0.02} & \textbf{0.86$\pm$0.00} \\
\end{tabular}
\caption{normalized Align score (nAS) and normalized Robinson Foulds score (nRF) on Test Set. Best results in bold. Models within confidence intervals of the best underlined. Uncertainty quanitified by 95\% confidence intervals based on 5 runs}
\label{table2}
\end{table}

We present the results in table \ref{table2}.
Among the three morphological representations tested, the mask-based model demonstrated the highest performance in phylogenetic trait extraction, achieving an average normalized Align Score (nAS) of 0.33$\pm$0.02 for the test set, significantly outperforming the Fourier-based model (0.45$\pm$0.01), though still within confidence intervals of the original model (0.39$\pm$0.07).

\section{Discussion}
It is unexpected that the mask-based model would outperform the full segmentation model which contains information about the external structures and shapes of the dorsal view of the beetle. We speculate if this indicates that those features are, in general, not as phylogenetically relevant as the shape. Further investigation would be necessary to confirm or disprove this theory. Perhaps this is also only true for this one taxonomic group, or perhaps these deep learning models are simply not yet advanced enough, or phylogenetically tuned enough, to focus on the phylogenetically relevant parts of the texture and structural information.

\begin{figure}[t!]
\centering
\includegraphics[width=.3\textwidth]{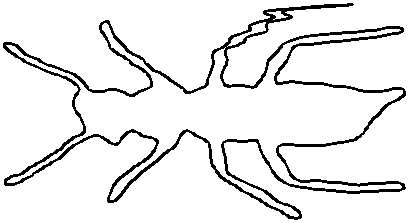}

\caption{Example of undesirable Fourier outline approximation which produced strange artifacts on the leg.}
\label{fig:badfourier}
\end{figure}

It is further unexpected that the Fourier-based model underperforms the mask-based model, given that Fourier descriptors are generally effective at capturing overall shape. We propose two plausible explanations for this outcome. First, as shown in table \ref{tab1:params}, the Fourier-based model operates with significantly fewer parameters compared to the ResNet50 architecture used for the binary mask, with the latter having over 200 times more parameters and incorporating greater depth and skip connections. This substantial reduction in model capacity likely limits the Fourier-based model's ability to extract complex or subtle features, potentially accounting for its lower performance. Second, while many Fourier descriptors effectively captured the beetles' outlines, we observed cases where the approximations exhibited unexpected distortions, particularly in finer structures such as the legs, as illustrated in figure \ref{fig:badfourier}. These inaccuracies likely diminished the utility of the Fourier descriptors for phylogenetic signal extraction, emphasizing the importance of ensuring high-fidelity approximations for reliable performance.

\section{Conclusion}
This study highlights the potential of deep learning to automate the extraction of phylogenetic traits from morphological data, particularly shape data, using rove beetles as an example. By comparing three distinct morphological representations—segmentations, binary masks and Fourier descriptors — we demonstrated that the mask-based model achieved the highest scores in phylogenetic inference. This finding underscores the utility of overall shape representations, which effectively capture phylogenetic signals while minimizing complexity. In contrast, Fourier descriptors and segmented features, while informative, faced limitations, however, also demonstrated that simpler models with fewer parameters may also achieve promising results. These results emphasize the importance of selecting appropriate morphological representations for automated phylogenetic analyses and pave the way for further integration of machine learning methods into evolutionary biology. Future work could explore the scalability of these approaches to larger datasets and more diverse taxa, as well as further investigate the phylogenetic importance of shape compared to texture and color in deep learning-based trait extraction.

\bibliography{main}

\end{document}